\documentclass[runningheads]{llncs}
\usepackage{graphicx}
\usepackage{tikz}
\usepackage{comment} 
\usepackage{amsmath,amssymb} % define this before the line numbering.
\usepackage{color}
\usepackage{hhline}
\usepackage{hyperref}
\usepackage{float}

\begin{document}
% \renewcommand\thelinenumber{\color[rgb]{0.2,0.5,0.8}\normalfont\sffamily\scriptsize\arabic{linenumber}\color[rgb]{0,0,0}}
% \renewcommand\makeLineNumber {\hss\thelinenumber\ \hspace{6mm} \rlap{\hskip\textwidth\ \hspace{6.5mm}\thelinenumber}}
% \linenumbers
\pagestyle{headings}
\mainmatter
\def\ECCVSubNumber{2966}  % Insert your submission number here

\title{An LSTM Approach to Temporal \\ 3D Object Detection in LiDAR Point Clouds}
% Replace with your title

% INITIAL SUBMISSION 
\begin{comment}
\titlerunning{ECCV-20 submission ID \ECCVSubNumber} 
\authorrunning{ECCV-20 submission ID \ECCVSubNumber} 
\author{Anonymous ECCV submission}
\institute{Paper ID \ECCVSubNumber}
\end{comment}
%******************

% CAMERA READY SUBMISSION
\titlerunning{An LSTM Approach to Temporal 3D Object Detection in Point Clouds}

\author{Rui Huang \and
Wanyue Zhang \and
Abhijit Kundu \and \\
Caroline Pantofaru \and 
David A Ross \and
Thomas Funkhouser \and
 Alireza Fathi
}
%{Frist Name: David, Last Name: A Ross}

\authorrunning{R. Huang et al.}
% First names are abbreviated in the running head.
% If there are more than two authors, 'et al.' is used.
%
\institute{Google Research \\
\email{huangrui@google.com}}
%******************
\maketitle

\begin{abstract}
Detecting objects in 3D LiDAR data is a core technology for autonomous driving and other robotics applications.  Although LiDAR data is acquired over time, most of the 3D object detection algorithms propose object bounding boxes independently for each frame and neglect the useful information available in the temporal domain. To address this problem, in this paper we propose a sparse LSTM-based multi-frame 3d object detection algorithm. We use a U-Net style 3D sparse convolution network to extract features for each frame's LiDAR point-cloud. These features are fed to the LSTM module together with the hidden and memory features from last frame to predict the 3d objects in the current frame as well as hidden and memory features that are passed to the next frame.  Experiments on the Waymo Open Dataset show that our algorithm outperforms the traditional frame by frame approach by 7.5\% mAP@0.7 and other multi-frame approaches by 1.2\% while using less memory and computation per frame. To the best of our knowledge, this is the first work to use an LSTM for 3D object detection in sparse point clouds.

\keywords{3D Object Detection, LSTM, Point Cloud}
\end{abstract}

\section{Introduction}
3D object detection is one of the fundamental tasks in computer vision.   Given observations of a scene with a 3D sensor (e.g., LiDAR), the goal is to output semantically labeled 3D oriented bounding boxes for all objects in every observation.   This task is critical for autonomous driving, object manipulation, augmented reality, and many other robot applications.  

Although almost all robot sensors capture data continuously (LiDAR, RGB-D video, RGB video, etc.), most 3D object detection algorithms consider only one ``frame'' of input sensor data when making bounding box predictions.  Historically, multi-frame data has not been widely available (e.g. the Kitti 3D Object Detection Challenge~\cite{geiger2012we} provides only one LiDAR sweep for each scene).  However, after datasets with multi-frame sequences of LiDAR were released~\cite{caesar2019nuscenes,chang2019argoverse,sun2019scalability}, most 3D object detection algorithms still work frame by frame. Among the algorithms with reported results on the nuScenes and Waymo object detection tasks, we find that only Ngiam et al.~\cite{ngiam2019starnet} and Hu et al.~\cite{hu20wysiwyg} consider multiple frames as input, and they both use simple methods based on reusing seed points or concatenating of input data for multiple frames. 

\begin{figure}
\centering
    \centering
    \includegraphics[width=.9\linewidth]{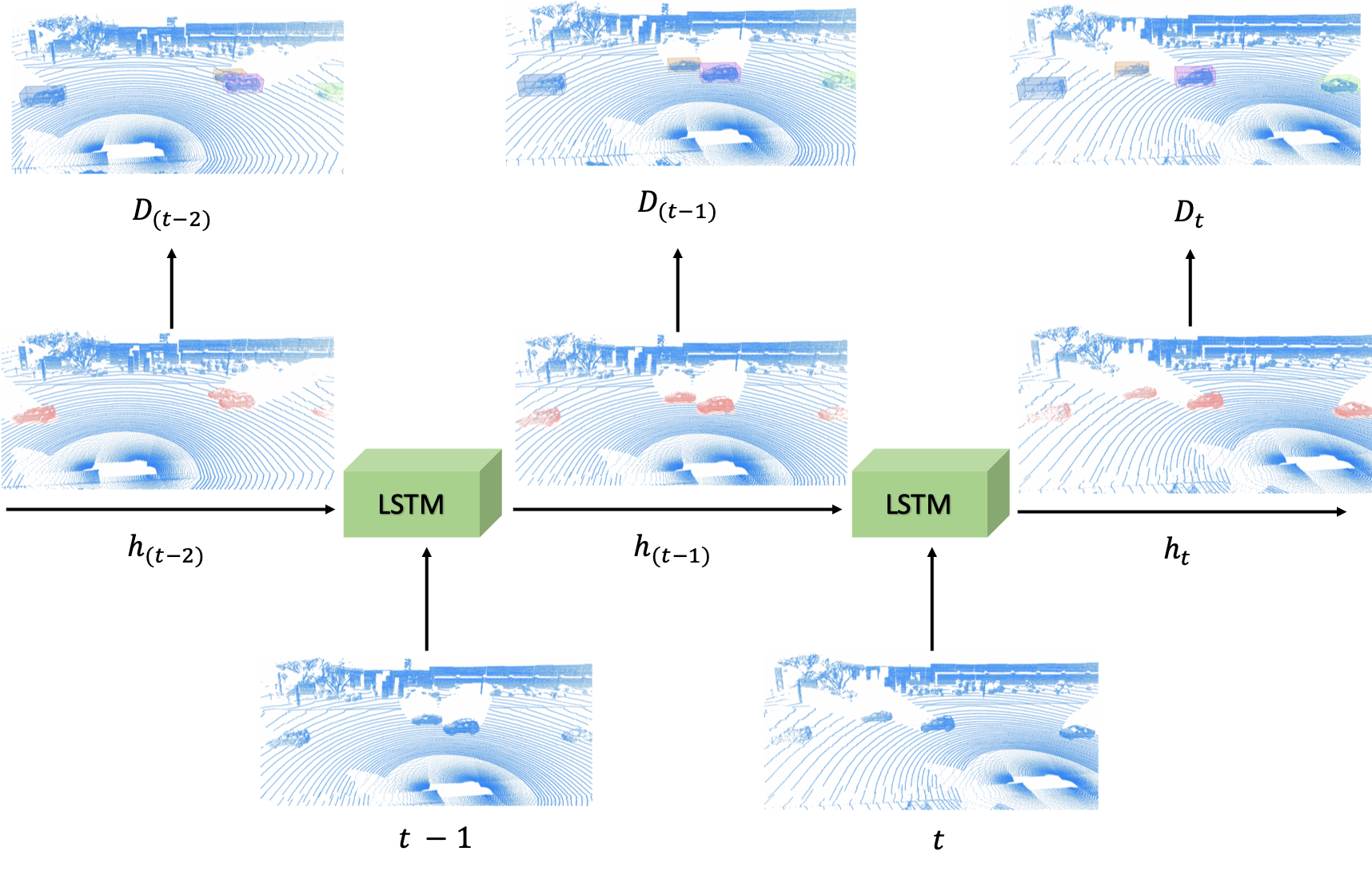}
    \caption{Our method consumes a sequence of point clouds as input. At each time step, the proposed LSTM module combines the point cloud features from the current frame with the hidden and memory features from the previous frame to predict the 3d objects in the current frame together with the hidden and memory features that are passed to the next frame. For memory efficiency we pass only the hidden feature points that have high score to the next frame (pink in the images). $D_t$ and $h_t$ represent the 3d object detections and hidden features in frame $t$ respectively.}
    \label{fig:thesis}
\end{figure}

In this paper, we investigate a new method as depicted in Fig~\ref{fig:thesis} that utilizes the temporal sequence of LiDAR data acquired by autonomous vehicle for 3D object detection.  Our approach is to use the memory of an LSTM to encode information about objects detected in previous frames in a way that can assist object detection in the current frame.  Specifically, we represent the memory and hidden state of the LSTM as 64-dimensional features associated with 3D points observed in previous frames.  At each frame, we use an LSTM architecture to combine features from these memory and hidden state 3D point clouds with features extracted from the latest observed 3D point cloud to produce bounding box predictions for the current frame and update the memory and hidden state for the next frame.  

The rationale for this approach is that the LSTM memory can represent everything known about object detections in the past in a concise set of features associated with a sparse set of 3D positions (ideally near past object detections).  In comparison to previous methods that concatenate input point clouds from multiple timesteps at every frame, this approach is more memory and compute efficient, as we include a relatively small number of 3D points related to past object detections in our LSTM memory rather than all the input points from previous frames (redundant storage and processing of 3D points on background objects such as trees and buildings is wasteful).  In comparison to a traditional LSTM, our approach of associating memory and hidden states with 3D points provides a spatial attention mechanism that assists object detection and enables transformation of the memory and hidden state from frame to frame based on the egomotion of the vehicle.  By associating the memory and hidden state with 3D points contributing to confident object detections in the past, we expect to get more accurate and robust detections with this approach.

Our implementation is built on a U-Net style sparse 3D convolution backbone (SparseConv) as described in~\cite{najibi2020dops}.  The point cloud for each input frame is voxelized into sparse 3d voxels, convolved on a sparse grid, and then associated with encoded features. Then, the encoded point cloud features are jointly voxelized with the memory and hidden features and passed through a SparseConv inside the LSTM. The LSTM network outputs hidden and memory point cloud features that will be passed to the next frame. Furthermore, the predicted hidden features are fed into the 3d detection head to generate per point object bounding box proposals (center, size, rotation, and confidence), which are further processed with a graph convolution to smooth per point predictions in local neighborhoods and non maximum suppression (NMS) to select a highly confident and non-overlapping set of proposed bounding boxes. The hidden and memory state (point features) only keep track of features in locations that have high objectness score. This enables us to be more memory efficient and to be able to aggregate and reason about information in a long sequence. 

Experiments show that this method outperforms frame by frame detection models by 7.5\% mAP@0.7 and beats a strong multi-frame concatenation baseline model by 1.2\%. Our model achieves 6.8\% better results than a baseline that refines predicted bounding boxes using the classical combination of frame by frame detection, Hungarian assignment, and Kalman filtering~\cite{Weng2019_3dmot}.

Our key contributions are summarized below: 
\begin{itemize}
\item We propose the first LSTM-based sequential point cloud processing framework for 3D object detection. It provides a significant performance boost over a single frame state-of-the-art 3D SparseConv model. Furthermore, our model outperforms a strong baseline based on concatenating multi-frame data.
\item We propose a 3D Sparse Conv LSTM where a small 3d sparse U-Net replaces the  fully connected layer in vanilla LSTM. Our model has explicit memory to facilitate reasoning across long sequence of point clouds. Compared to point-based methods, our voxel-based module is effective and efficient in fusing accumulated memory and input data in multiple scales, while maintaining a constant memory footprint regardless of the sequence length in inference time. 
\end{itemize}

\section{Related Work}
\subsubsection{3D Object Detection}

A common approach to 3D object detection is to utilize ideas that have been successful for 2D object detection~\cite{yang2018pixor,simon2018complex,simon2019complexer,chen2017multi,liang2018deep,li2016vehicle,meyer2019lasernet}. For instance, Frustum-PointNet~\cite{qi2018frustum} uses 2D detectors on RGB images and point clouds from the depth sensor. However, the search space for potential objects is limited in the 3D viewing frustum extended from 2D regions. MV3D~\cite{chen2017multi} deploys a multi-view fusion network for features extracted from the bird-eye view, Lidar range view and RGB images. Building on soft voxelization, Zhou et al.~\cite{zhou2019end} fuse features based on Cartesian coordinate, perspective coordinate and the output of a shared fully connected layer from LiDAR points.

Another class of methods~\cite{yang2018foldingnet,zhao20193d,groueix2018atlasnet,zhao2019pointweb,shi2019pointrcnn,qi2019deep,qi2017pointnet++} propose networks that directly consume the 3d point cloud as input. 
Shi et al.~\cite{shi2019pointrcnn} propose a bottom-up approach to directly generate 3D bounding box proposals from the point cloud, followed by a sub-network for refinement. VoteNet~\cite{qi2019deep} uses PointNet++~\cite{qi2017pointnet++} backbone to vote for object centers. The votes are then clustered to produce the final bounding box proposals.

There is an increasing trend to convert point clouds to regular grids where 3d convolution can be conveniently constructed. VoxelNet~\cite{zhou2018voxelnet} partitions point clouds into voxels but this method is computationally expensive. Some of the previous works attempt to solve this issue by making use of the sparsity pattern of 3D points~\cite{engelcke2017vote3deep,sparseconvs2015,graham20183d,riegler2017octnet,engelcke2017vote3deep,najibi2020dops}. SparseConv~\cite{SubmanifoldSparseConvNet,najibi2020dops} is exceptionally efficient as convolutions are restricted to active sites and sparsity pattern is preserved even after layers of convolutions. Our work uses a sparse voxel U-Net as the backbone as described in~\cite{najibi2020dops}. 

\subsubsection{Spatio-temporal Methods}
Various ways to make use of the temporal information are experimented for different vision tasks such as prediction in video data and modeling the human motion dynamics~\cite{xu2019unsupervised,xiao2018video,chen2019temporally,huang2019attention,humanMotionKanazawa19}. In addition, there are various LSTM based methods for object detection in video~\cite{feng2019spatio,kang2017object,xiao2018video,chen2019temporally}. Among those, Xiao et al.~\cite{xiao2018video} introduce a spatio-temporal memory module (STMM) to model temporal appearance and motion changes of objects. Teng et al.~\cite{teng2018clickbait} explore detecting objects in streaming video using weak supervision by tracking and optical flow. 

For LiDAR point clouds, ~\cite{Yin_2020_CVPR,el2018yolo4d,mccrae3d} use ConvGRU or ConvLSTM to process the bird-eye view projection. Luo et al.~\cite{luo2018fast} explore the synergy of 4 tasks for autonomous driving: detection, tracking, motion forecasting and motion planning. By concatenating multiple frames of input, 2d convolution is performed on voxels to forecast the next $n$ frames. Inspired by Luo et al.~\cite{luo2018fast}, Casas et al.~\cite{casas2018intentnet} jointly tackle detection and motion estimation by adding the rasterized map to provide the environment context for more accurate forecasting.  MeteorNet~\cite{liu2019meteornet} processes point clouds directly and proposes direct grouping and chained grouping to find nearest spatio-temporal neighborhours. This method is applied to semantic segmentation, classification and flow estimation. Choy et al.~\cite{choy20194d} augment 3D data with the time axis and build sparse 4D convolutions using non-conventional kernel shapes. Our approach is distinct from the above as we propose a 3d sparse LSTM model that consumes sparse 3d data and performs 3d sparse operations and merging to perform 3d object detection.

The closest related work to ours is PointRNN~\cite{fan2019pointrnn}, which adapts RNNs for predicting scene flow on multi-frame point clouds. It proposes a point-rnn function to aggregate the past state and the current input based on the point coordinates. Our approach is different as we conduct 3D sparse convolution on adjacent voxels which avoids the expensive step of finding nearest neighbors for each point in the point cloud. Besides, no permutation invariant aggregation is needed by our method. Furthermore, we focus on 3d object detection in a sequence while PointRNN~\cite{fan2019pointrnn} focuses on scene flow.

\subsubsection{3D Scene Flow and Object Tracking}

Some researchers have focused on the related problem of predicting scene flow (3D motion vector per point) from pairs of input point clouds in adjacent frames.   Since the magnitude and direction of movement of points provides a cue for object detection, these two tasks could provide mutual context for each other. For instance, Behl et al.~\cite{behl2017bounding} extract xyz object coordinates from 4 RGB images and incorporate detection and instance segmentation cues from 2D to improve scene flow. PointFlowNet~\cite{behl2019pointflownet} gets rid of the reliance on 2D images by using an encoder-decoder model to tackle flow, object location and motion in conjunction. FlowNet3D~\cite{dosovitskiy2015flownet} consumes point clouds directly by using a Set Conv Layer to down-sample points, a flow embedding layer to aggregate features from two point clouds and a Set UpConv layer to get a per-point estimation of the translation vectors.  While these methods are loosely related to ours, they are aimed at predicting flow for the entire scene and do not aim at improving 3d object detection.

Some of the previous works~\cite{Weng2019_3dmot,chiu2020probabilistic,chang2019argoverse} have focused on 3D tracking of objects. However, these algorithms mainly focus on generating multi-frame tracks and do not necessarily result in a more accurate per frame 3d object detection. Wang et al.~\cite{Weng2019_3dmot} detect objects in every frame and then use the Hungarian algorithm to associate objects and Kalman filter to aggregate predictions across frames. We use this method as a baseline and compare our 3d object detection accuracy with theirs in the results section.

\section{Method}
\begin{figure*}[!ht]
\centering
\includegraphics[width=0.95\linewidth]{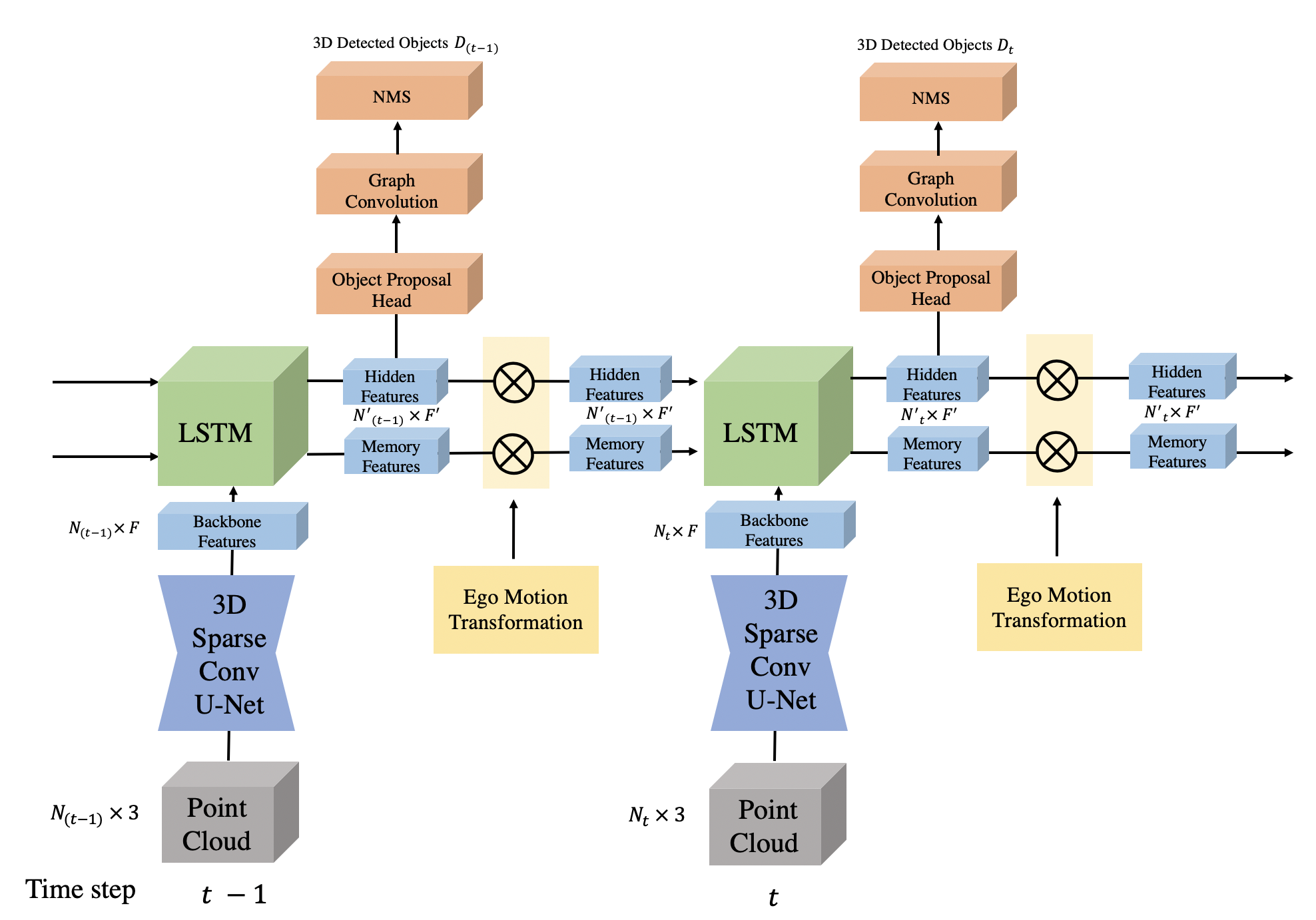}
\caption{Overview of the temporal detection framework: A sequence of point clouds are  processed by a Sparse Conv U-Net backbone in each frame. The 3d sparse LSTM fuses the backbone feature at the current time step $t$ with the hidden and memory feature at the previous time step $t-1$ to produce hidden and memory feature at time step $t$. Object proposals are generated from the hidden feature and refined using a graph convolution network. Farthest point sampling and non-maximum suppression are applied to the proposed 3d objects to produce the final detected 3d objects.}
\label{fig:architecture}
\end{figure*}

The architecture of our method is shown in Fig \ref{fig:architecture}. In each frame, we extract point features from the input point cloud by feeding it into a 3d sparse voxel conv U-Net as described in~\cite{najibi2020dops}. We feed the extracted features together with the  memory and hidden features from previous frame to our proposed 3d sparse conv LSTM which processes them and outputs the hidden and memory features that will be consumed by the next frame. In the mean time, an object detection head is applied to the hidden features to produce 3d object proposals in each frame. The proposals are then passed through a graph convolution stage and then non-maximum suppression to output the detected 3d objects for each frame.

\subsection{3D Sparse Conv U-Net}
To extract point features from the input point cloud, we use a U-Net shaped backbone as described in~\cite{najibi2020dops}. The input to our feature extractor is a point cloud as a $N\times3$ tensor (points with their xyz position). The network first voxelizes the point cloud into sparse 3d voxels. If multiple points fall in the same voxel, the voxel feature would be the average of the xyz location of those points. 
The netowrk encoder consists of several blocks of 3d sparse convolution layers where each block is followed by a 3d max pooling. The U-Net decoder upsamples the spatial resolution gradually with several blocks of sparse convolution layers and upsampling with skip connections from the  encoder layers. The extracted voxel features are de-voxelized back to the points to output a $N \times F$ tensor where $F$ is the extracted feature dimension.

\subsection{3D Sparse Conv LSTM}
\begin{figure*}[!ht]
\centering
\includegraphics[width=0.7\linewidth]{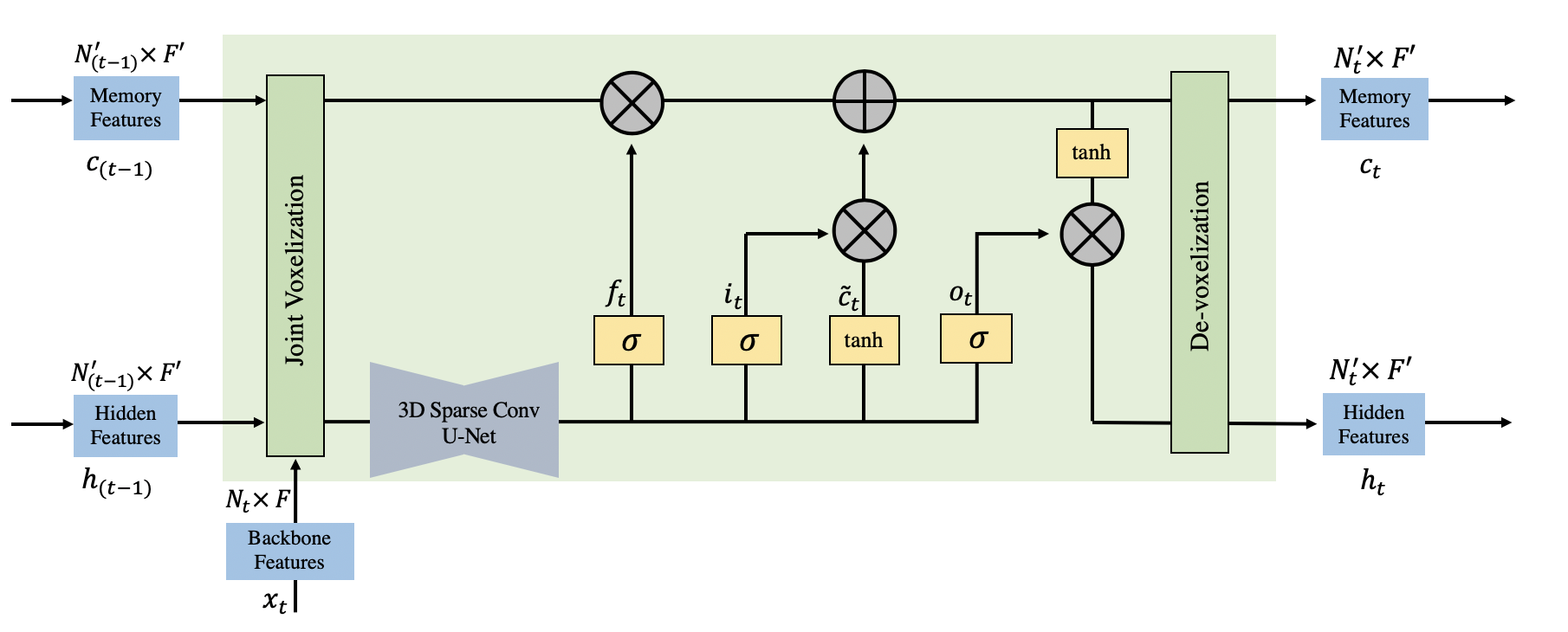}
\caption{3D sparse conv LSTM structure. The backbone feature $x_{t}$, memory feature $c_{t-1}$ and hidden feature $h_{t-1}$ are jointly voxelized. A lightweight SparseConv U-Net takes the concatenation of $x_{t}$ and $h_{t-1}$ to produce gates and memory candidate. Output features from the LSTM are de-voxelized before being sent to the next time step.}
\label{fig:lstm}
\end{figure*}

We use an LSTM based on 3d sparse conv network to leverage the temporal information in the sequence of LiDAR frames. Here we first review the basic notation for our 3d sparse LSTM module (Fig \ref{fig:lstm}) and then introduce the key differences and challenges in more details.

\textbf{Vanilla LSTM}:
Long short term  memory (LSTM)~\cite{hochreiter1997long} is a common variant of recurrent neural network used extensively for time series data and natural language processing. The vanilla LSTM structure is described below:

\begin{equation}
    f_t = \sigma(W_f \cdot [h_{t-1}, x_t] + b_f) \label{eq:2}
\end{equation}
\begin{equation}
    i_t = \sigma(W_i \cdot[h_{t-1}, x_t] + b_i) \label{eq:3}
\end{equation}
\begin{equation}
    \tilde{c}_t = tanh(W_c \cdot [h_{t-1}, x_t] + b_c) \label{eq:4}
\end{equation}
\begin{equation}
    c_t = f_t \times c_{t-1} + i_t \times \tilde{c}_t \label{eq:5}
\end{equation}
\begin{equation}
    o_t = \sigma(W_o[h_{t-1}, x_t] + b_o) \label{eq:6}
\end{equation}
\begin{equation}
    h_t = o_t \times tanh(c_t) \label{eq:7}
\end{equation}
The input feature at current time step $x_t$ and the hidden feature at the previous time step $h_{t-1}$ are concatenated before being transformed by a fully connected layer with weight matrix  $W$ and bias $b$. The transformed feature is activated by either sigmoid ($\sigma$) or tanh function to produce input gate ($i_t$), forget gate ($f_t$), output gate ($o_t$) and cell memory candidate ($\tilde{c}_t$) for the current time step.
The cell memory $c_t$ is updated from $\tilde{c}_t$ and the cell memory at previous time step $c_{t-1}$, where $\times$ denotes element-wise multiplication.

\textbf{LSTM on Sparse Point Clouds}: In our context, $x_t$ of size $N_{t} \times F$ is the point cloud features that is extracted using our 3d sparse backbone, $h_{t-1}$ and $c_{t-1}$ of size $N'_{t-1} \times F'$ are hidden and memory point features respectively. 
We subsample the hidden and memory point features and only keep a subset ($N'_{t-1}$) of the points that have high semantic scores (obtained from the pre-trained single frame detection model).

In order for the LSTM to be able to fuse multiple 3d sparse tensor features, we replace the fully connected layer in vanilla LSTM with a lightweight 3d sparse conv U-Net structure to produce gates and cell memory candidate. This approach ensures that the LSTM has enough capacity to conduct sequential reasoning in the 3d sparse space and avoid the expensive nearest neighbor search in point-based methods~\cite{fan2019pointrnn,liu2019flownet3d,liu2019meteornet}.

\textbf{Joint Voxelization}:
Due to the object motion in the scene (even though that we compensate for the egomotion), $x_t$ and $h_{t-1}$ (or $c_{t-1}$) will not align in the 3D space.
Our solution to this problem is to jointly voxelize the three point clouds, namely $x_t$, $h_{t-1}$ and $c_{t-1}$. The resulting three voxel grids are then concatenated in feature dimension. If one of the sparse point features has no point in a voxel but the other ones do, we will pad the voxel features of the one with the missing point with zeros.

In other words, since the point clouds are non-overlapping in some region, after joint voxelization there will be empty voxels inserted into each voxel grids in non-overlapping region. This means that the joint sparse voxel representation covers the union of the spatial extent of all participating point clouds which is still extremely sparse.

\subsection{Object Detection}

The proposal head takes the voxelized hidden features $h_{t}$ of size $N \times F'$ from the LSTM at each time step to independently generate per voxel bounding box proposals (center, rotation, height, length and width). The predictions are then de-voxelized to produce per point bounding box predictions, taking into account each point's position offset within the voxel. During de-voxelization we transfer the prediction associated with each voxel to all the points that fall inside it. The head is implemented with 3 layers of sparse convolutions for each attribute.

As described in~\cite{najibi2020dops}, we construct a graph on top of the per point predictions, where each point (node) is connected to its K nearest neighbors with similar object center predictions. The predicted object attributes are propagated based on a predicted weight per point. The weight determines the significance of each point in comparison with its neighbors. The bounding box prediction loss is applied both before and after the propagation.

During the inference time, we sample a subset of high score and farthest predictions and then apply non-maximum suppression to output the final 3d object detection results.

\subsection{Training}

We first train a single frame backbone which we use to extract the encoded point features for each frame. The proposed LSTM module processes the encoded point features $x_{t}$ together with hidden and memory point features from previous frame ($h_{t-1}$ and $c_{t-1}$) and outputs hidden and memory features $h_{t}$ and $c_{t}$ for the current frame.
The object detection head takes $h_{t}$ as input and outputs the 3d detected objects in frame $t$. The 3d box regression and classification losses are applied to the outputs in every frame.
Our algorithm operates in the local coordinate frame which means the features from previous frames are transformed to the current frame to compensate for egomotion.

As described in~\cite{najibi2020dops}, we adopt a hybrid of regression and classification losses for bounding box prediction. Each bounding box is represented by height, length, width, center location and a $3 \times 3$ rotation matrix. Instead of computing a separate loss for each of the parameters, we use an integrated box corner loss which can be back propagated and get all the individual attributes updated at once. We first calculate the 8 box corners in a differentiable way, then apply Huber loss on the distance between ground-truth and predicted boxes. The benefit of doing this is the ease of training as we do not have to tune multiple individual losses. 

We use a dynamic classification loss as described in~\cite{najibi2020dops}. At each step, we classify the predictions that have more than $70\%$ IOU with their corresponding ground-truth object as positive and the rest of the predictions as classified as negative. As the model gets better in box prediction, there will be more positive predicted boxes over time.

\section{Experimental Results}
\label{sec:results}

We perform a series of experiments to evaluate the performance of our LSTM network in comparison to the alternative approaches. Furthermore, we study the effects of our design decisions through ablation studies. 

\begin{figure}[ht]
\centering
    \centering
    \includegraphics[width=1.0\linewidth]{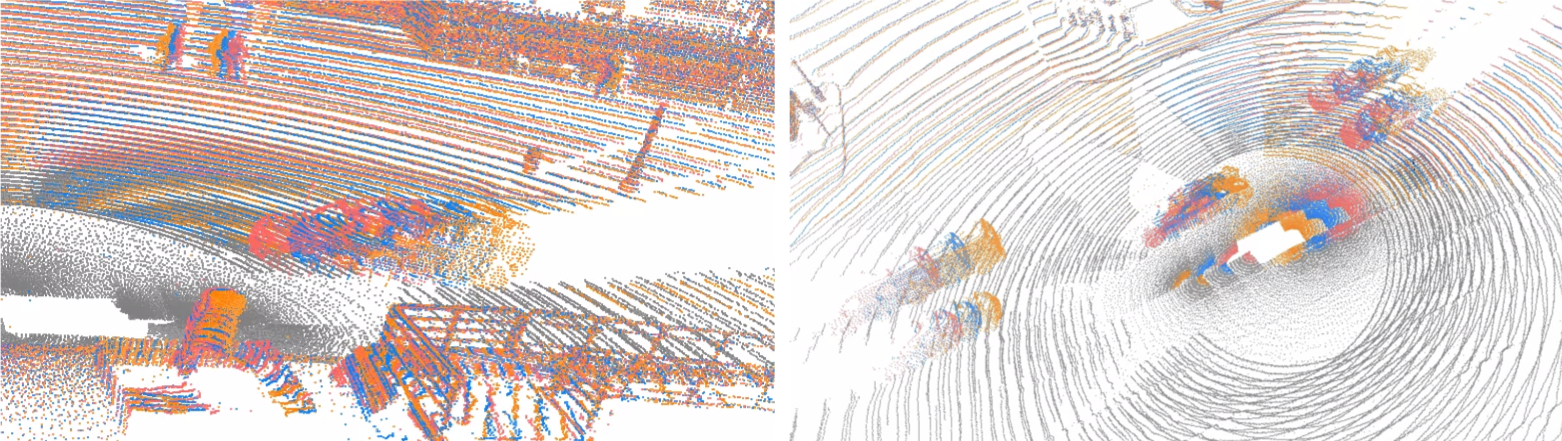}
    \caption{Example sequences in Waymo Open Dataset. Each frame is colored differently. A few fast moving cars and two walking pedestrians are shown over three frames.}
    \label{fig:scene_sequence}
\end{figure}

\textbf{Dataset}:
We use the recently released Waymo Open Dataset~\cite{sun2019scalability} for our experiments. It contains 1000 sequences (798 training and 202 validation) captured in major US cities under diverse weather conditions and times of the day. 
Each sequence (Fig \ref{fig:scene_sequence}) has approximately 200 frames at a frame rate of 100ms. Each frame has multiple LiDARs and cameras with annotated 3D and 2D bounding box labels for vehicles, pedestrians, cyclists, signs. In our experiments, only LiDAR point clouds and 3D bounding box labels for vehicles with 5+ points are used for training and evaluation. The Waymo Open Dataset is a larger scale dataset in comparison to the previous self-driving car datasets such as Kitti dataset~\cite{geiger2012we}. The 20 seconds sequence for each scene enables the training and evaluation of the temporal 3D object detection task on point clouds in challenging and realistic autonomous driving scenarios.

\begin{table}[ht]
\small
\setlength{\tabcolsep}{3.2pt}
\begin{center}
\begin{tabular}{|c|c|c|}
    \hline
    Model & mAP@0.7IoU \\  \hline
    StarNet~\cite{ngiam2019starnet} & 53.7\\ \hline
    PointPillars$\dag$~\cite{lang2019pointpillars} & 57.2\\ \hline
    MVF~\cite{zhou2019end}  & 62.9\\ \hhline{|=|=|}
    U-Net &56.1\\ \hline
    U-Net + Kalman Filter~\cite{Weng2019_3dmot}&56.8 \\ \hline
    Concatenation (4 frames) & 62.4 \\ \hline
    Ours (4 frames) & \textbf{63.6}\\ 
    \hline
\end{tabular}
\end{center}
%   \vspace{1cm}
\caption{3D object detection results on Waymo Open dataset \textit{validation} set. Unless noted otherwise, the models are using single frame. $\dag$:re-implemented by ~\cite{ngiam2019starnet}.} 
\label{tab:waymo}
\end{table}

\textbf{Experiment Details}:
We follow the metric used in almost all self-driving car datasets which is the mean average precision (mAP) metric for the 7 degree-of-freedom 3D boxes at intersection over union (IoU) threshold of 0.7 for vehicles.

For the object detection backbone, the encoder contains 6 blocks each with two 3D SparseConv layers, with output feature dimensions of 64, 96, 128, 160, 192, 224, 256.
The decoder has the same structure in reversed order with skip connections from encoder layers.

We use a lightweight 3D sparse U-Net for LSTM that has one encoder block (of 128 dimensions), max pooling, one bottleneck block (of 128 dimensions), unpooling, and one decoder block (of 256 dimensions).
Models are trained on 20 synced GPUs with a batch size of 2 (which means effectively a batch size of 40).
We train the model with 0.1 initial learning rate. After 25k steps, we decay the learning rate every 7k steps by the factors of [0.3, 0.1, 0.01, 0.001, 0.0001]. 
We use a voxel size of [0.2m, 0.2m, 0.2m]. The LSTM module uses the sub-sampled point cloud features that are computed by the backbone as well as hidden and memory features that it receives from the previous frame. Hidden feature points in previous steps are accumulated during the sequence. In practice, this only slightly increase the number of non-empty voxels at each step.

\begin{figure}[h]
\centering
    \centering
    \includegraphics[width=1\linewidth]{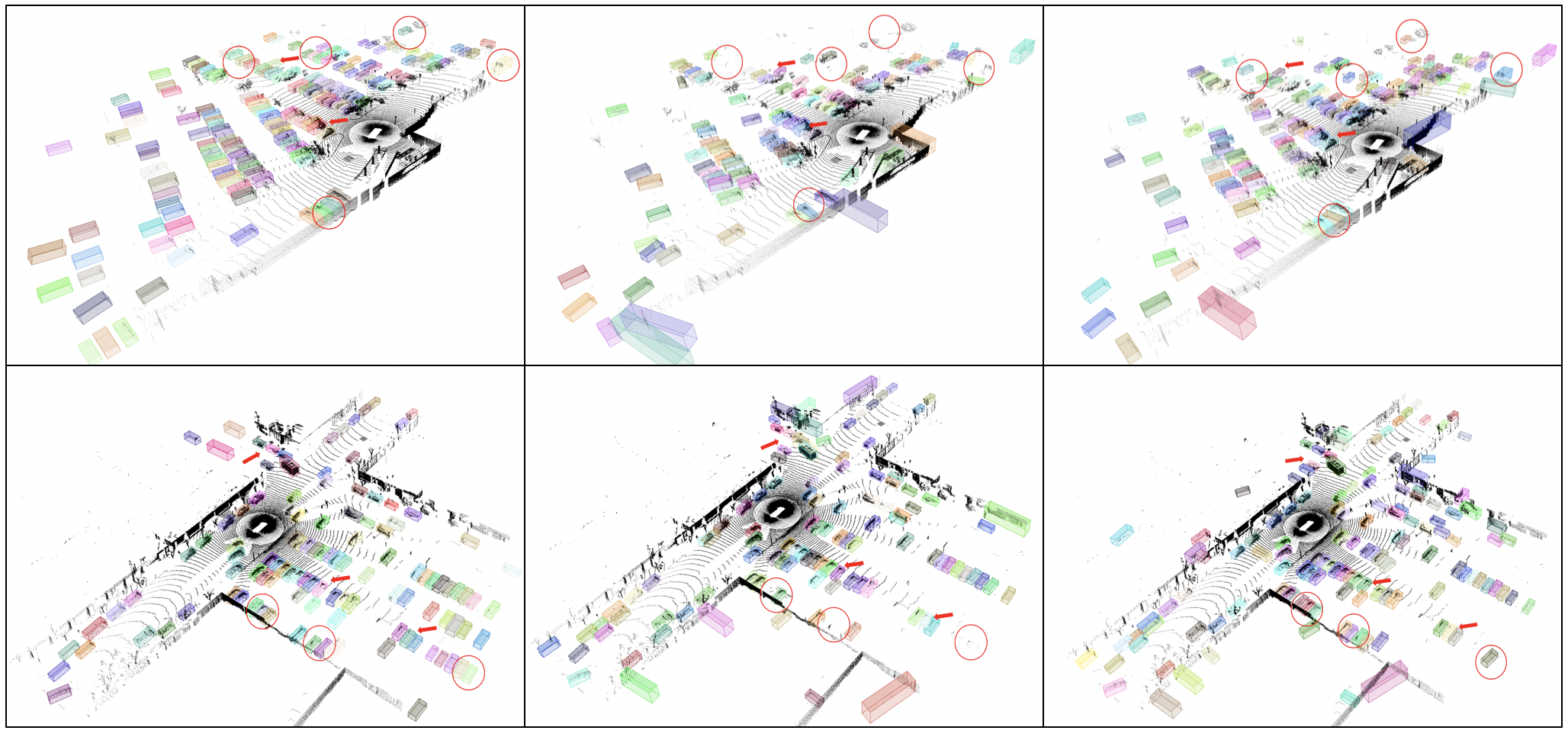}
    \caption{We compare our sparse LSTM 3d object detection results with the one frame 3d detection baseline. Left: ground truth labels; Middle: single frame predictions; Right: LSTM predictions. Misaligned (arrows) and missing (circles) vehicles are highlighted.}
    \label{fig:qualitative_results}
\end{figure}

\subsection{Object Detection Results}

We show our results on Waymo Open Dataset in Table~\ref{tab:waymo}. Our first baseline (U-Net) is a single frame model built on our sparse 3D convolution U-Net backbone (without the LSTM) which achieves 56.1\% mAP at IoU 0.7. Our second baseline combines the single frame detector with AB3DMOT~\cite{Weng2019_3dmot}, which deploys a combination of 3D Kalman Filter and Hungarian algorithm. Kalman filter is a classical way for tracking objects which we use to  update measurements based on prior from previous frames\footnote{Using code released by ~\cite{Weng2019_3dmot}.}. We build this baseline by applying the Kalman filter on top of the single frame detector. Based on our experiments, this method achieves 0.7\% gain in comparison to the single frame baseline.

Our third baseline feeds the concatenation of 4 frames into our U-Net backbone (same as the first baseline, but with 4 frames of input).  We concatenate the points in the feature dimension after applying ego-motion transformation. Since points in different frames do not align, we use zero padding to offset the features. This is more flexible than Luo et al.~\cite{luo2018fast}'s early fusion with 1D convolution, and more memory and compute efficient than their late fusion since the backbone runs only once. In comparison to the U-Net baseline, this gives rise to a 6.3\% increase of mAP to 62.4\%.  Finally, in the last row we show our proposed LSTM model (4-frames) results with the best performance of 63.6\% mAP@0.7.

We report the results of other single-frame detectors for comparison. StarNet~\cite{ngiam2019starnet} is a point-based detector based on sampling instead of learned proposals. It achieves 53.7\% on the validation dataset. PointPillars~\cite{lang2019pointpillars} organizes point clouds into regular vertical columns and detects objects using 2D CNN. It achieves 57.2\% mAP (re-implemented by~\cite{ngiam2019starnet}). MVF~\cite{zhou2019end} has the state-of-the-art single frame results on the Waymo Open Dataset. However, their method is not directly comparable to ours since they perform significant data augmentation.  Regardless, our focus is on how an LSTM can be used to improve the results of any method, which is largely orthogonal to any particular choice of single-frame baseline. The results demonstrate its effectiveness (7.5\% improvement in mAP@0.7 over the single frame model with the same U-Net backbone).

\begin{figure}[H]
\centering
    \centering
    \includegraphics[width=1.0\linewidth]{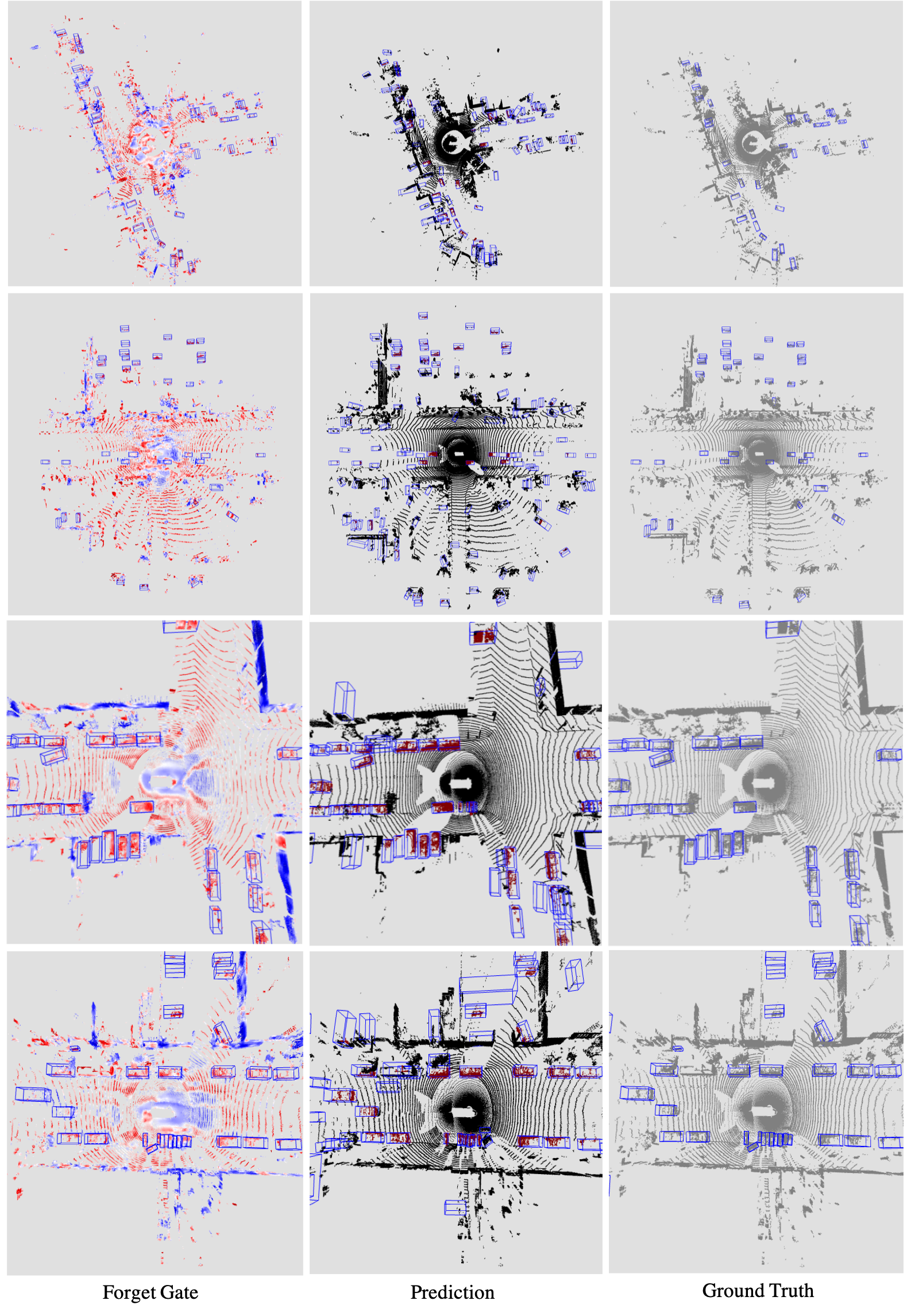}
    \caption{Visualization of the forget gate of our proposed LSTM module, prediction of object detection and the ground truth label. 
    The gate is visualized as a heatmap, where a high value (red) intuitively means \textit{not} forgetting the hidden features at the location. 
    We included ground truth boxes in this figure for clarity.
    The points within vehicles are mostly in high value, while the buildings and other objects are mostly in blue.
    The color in the prediction represents semantic classes of vehicles (red), background (black). }
    \label{fig:prediciton}
\end{figure}

The qualitative result of our method is shown in Fig~\ref{fig:qualitative_results} and Fig~\ref{fig:prediciton}. 
Fig~\ref{fig:qualitative_results} shows that our LSTM method predicts more accurate bounding boxes and have fewer false negatives in comparison to the single frame baseline.

\begin{figure}[H]
\centering
    \centering
    \includegraphics[width=.95\linewidth]{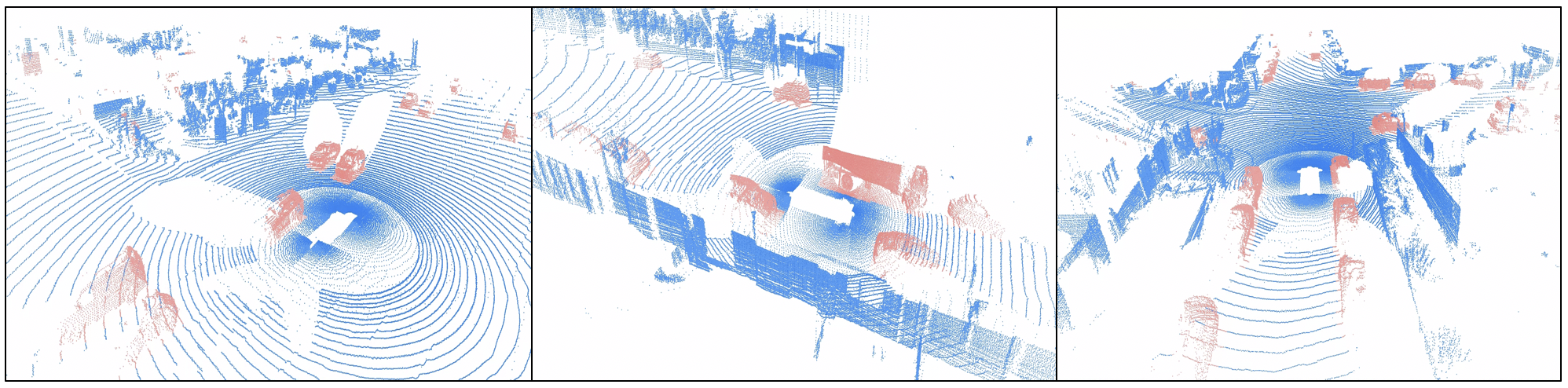}
    \caption{Locations where hidden and memory features are selected (pink points).}
    \label{fig:hidden_state}
\end{figure}

Due to the NMS process, there are often more predicted boxes than in the ground truth.
These false positives usually have low semantics scores (confidence).
 For better visualization, the forget gate feature heat maps (Fig \ref{fig:prediciton}) are sampled in point locations of a full point cloud from a voxel grid.
 The actual memory features point cloud (pink in Fig~\ref{fig:hidden_state}) concentrates on a smaller spatial extend, mostly on object surfaces. The memory features indicate the spatial attention of the network, which is useful to carry the most relevant information from previous frames to future frames.

In Table~\ref{tab:ablation}, we present the results of our LSTM model with different number of frames. In the first row, as a sanity check, we show the mAP accuracy when applying our LSTM model to one frame. We see a 2.6\% increase in comparison to the one frame raw U-Net model shown in row 3 of Table~\ref{tab:waymo}. This is because our LSTM model has convolutional and non-linear layers and gates that enhance the expressiveness of the network. We see a 1.0\% improvement when using a 2-frame LSTM model in comparison to the 1-frame one. For the 4-frame LSTM model, mAP reaches 63.6\%, with a 4.9\% improvement in comparison to the 1-frame LSTM model.

In order to take 7 frames as input, we decrease the batch size from 2 to 1 due to memory constraints. Compared with the 4 frame model with batch size 1, the performance increases by 1.0\%. Overall, the ablation study shows that LSTM with hidden and memory features over longer sequences results in higher 3d object detection accuracy.

\begin{table}[h]
\small
\setlength{\tabcolsep}{3.2pt}
\begin{center}
\begin{tabular}{|c|c|}
    \hline
    Model & mAP@0.7IoU\\ \hline    

    1 frame & 58.7 \\ \hline
    2 frames & 59.7 \\ \hline
    4 frames &\textbf{63.6}\\ \hhline{|=|=|} 
    4 frames (batch size 1)& {62.3}\\ \hline
    7 frames (batch size 1)& {63.3}\\

    \hline
\end{tabular}
\end{center}
%   \vspace{1cm}
\caption{Ablation studies of our detection model on Waymo Open \textit{validation} set} 
\label{tab:ablation}
\end{table}

\section{Discussion}

We have proposed an LSTM approach for detecting 3D objects in a sequence of LiDAR point cloud observations.  Our method leverages memory and hidden state features associated with 3D points from previous object detections, which are transformed according to vehicle egomotion at each timestep.   The backbone for our LSTM is a sparse 3D convolution network that co-voxelizes the input point cloud, memory, and hidden state at each frame. Experiments on Waymo Open Dataset demonstrate that our algorithm achieves the state-of-the-art results and outperforms a single frame baseline by 7.5\%, a multi-frame object detection baseline by 1.2\%, and a multi-frame object tracking baseline by 6.8\%.  In the future, we would like to also predict scene flow and use it to better transform the memory and hidden states in our LSTM, and we would like to study how our LSTM can be used to improve a variety of other single-frame object detectors.

\textbf{Memory Efficiency}:
Our proposed model is more memory efficient in comparison to previous temporal models that concatenate the point clouds from multiple frames~\cite{lang2019pointpillars,luo2018fast}.   A method that concatenates $M$ frames needs to apply the 3d network at each frame to $M$ times more number of points, while our 3d network is only applied to the points in the current frame plus a small set of features coming from last frame. Please note that our sub-sampled hidden and memory feature points (we sample 30k points in each frame out of 180k LiDAR points) are lightweight in comparison to passing full size point features from previous frames.

\textbf{Computation Efficiency}:
In comparison to the single frame model, the LSTM module adds a very small computational overhead. LSTM runs in a stream and is able to reuse the intermediate tensors that are computed in the previous time steps. The only additional overhead to per-frame computation is 3 sparse conv blocks which is a small fraction (10\% of the parameters) of the single frame network that uses 15 sparse conv blocks. Note that our single frame feature extractor runs in 19ms on a Titan V GPU. Given that the lidar input arrives at 10hz, our network is still able to run in real-time within its 100ms computation budget. Therefore, it adds only a small overhead while it gains 7.5\% in comparison to our single frame method. In comparison to the 4-frame concatenation baseline, the LSTM approach is more efficient. Concatenation reduces the sparsity and results in feeding a denser set of voxels to the network. We show that not only LSTM method is more efficient but also it achieves 1.2\% better result.

\par\vfill\par

\clearpage
% ---- Bibliography ----
%
% BibTeX users should specify bibliography style 'splncs04'.
% References will then be sorted and formatted in the correct style.
%
\bibliographystyle{splncs04}
\bibliography{egbib}
\end{document}